\documentclass[runningheads]{llncs}
\usepackage{graphicx}
\usepackage{amsmath,amssymb} 
\usepackage{color}

\begin{document}
\title{Non-rigid 3D Shape Registration using an Adaptive Template} 

\titlerunning{Non-rigid 3D Shape Registration using an Adaptive Template}
\author{Hang Dai, Nick Pears and William Smith}
\authorrunning{Hang Dai, Nick Pears and William Smith}
\institute{Department of Computer Science, University of York, UK \\
\email{\{hd816,nick.pears,william.smith\}@york.ac.uk}}

\maketitle

\begin{abstract}
We present a new fully-automatic non-rigid 3D shape registration (morphing) framework comprising (1) a new 3D landmarking and pose normalisation method; (2) an adaptive shape template method to improve the convergence of registration algorithms and achieve a better final shape correspondence and (3) a new iterative registration method that combines Iterative Closest Points with Coherent Point Drift (CPD) to achieve a more stable and accurate correspondence establishment than standard CPD. We call this new morphing approach \emph{Iterative Coherent Point Drift} (ICPD). Our proposed framework is evaluated qualitatively and quantitatively on three datasets: Headspace, BU3D and a synthetic LSFM dataset, and is compared with several other methods. The proposed framework is shown to give state-of-the-art performance.

\end{abstract}

\section{Introduction}

The goal of non-rigid shape registration is to align and morph a \emph{source} point set to a \emph{target} point set. By using some form of template shape as the source, morphing is able to reparametrise a collection of raw 3D scans of some object class into a consistent form. This facilitates full dataset alignment and subsequent 3D Morphable Model (3DMM) construction. In turn, the 3DMM constitutes a useful shape prior in many computer vision tasks, such as recognition and missing parts reconstruction.

Currently, methods that deform a 3D template to all members of a specific 3D object class in a dataset use the same template shape. However, datasets representative of global object classes often have a wide variation in terms of the spatial distribution of their constituent parts. Our object class in this paper is that of the human face/head, where the relative positions of key parts, such as the ears, mouth, and nose are highly varied, particularly when trying to build 3DMMs across a wide demographic range of age, gender and ethnicity. Using a single template shape means that often key parts of the template are not at the same relative positions as those of the raw 3D scan. This causes slow convergence of shape morphing and, worse still, leads to end results that have visible residual errors and inaccurate correspondences in salient local parts. 

To counter this, we propose an adaptive template approach that provides an automatically tailored template for each raw 3D scan in the dataset. The adaptive template is obtained from the original template using \emph{sparse} shape information (typically point landmarks), thereby locally matching the raw 3D scan very specifically. Although this is a pre-process that involves template shape adaptation, we do not consider it as part of the main template morphing process, which operates over \emph{dense} shape information. 

\begin{figure}[t!]
\begin{center}
\includegraphics[clip, trim=0.5cm 1cm 0.5cm 0.6cm,width=1\linewidth]{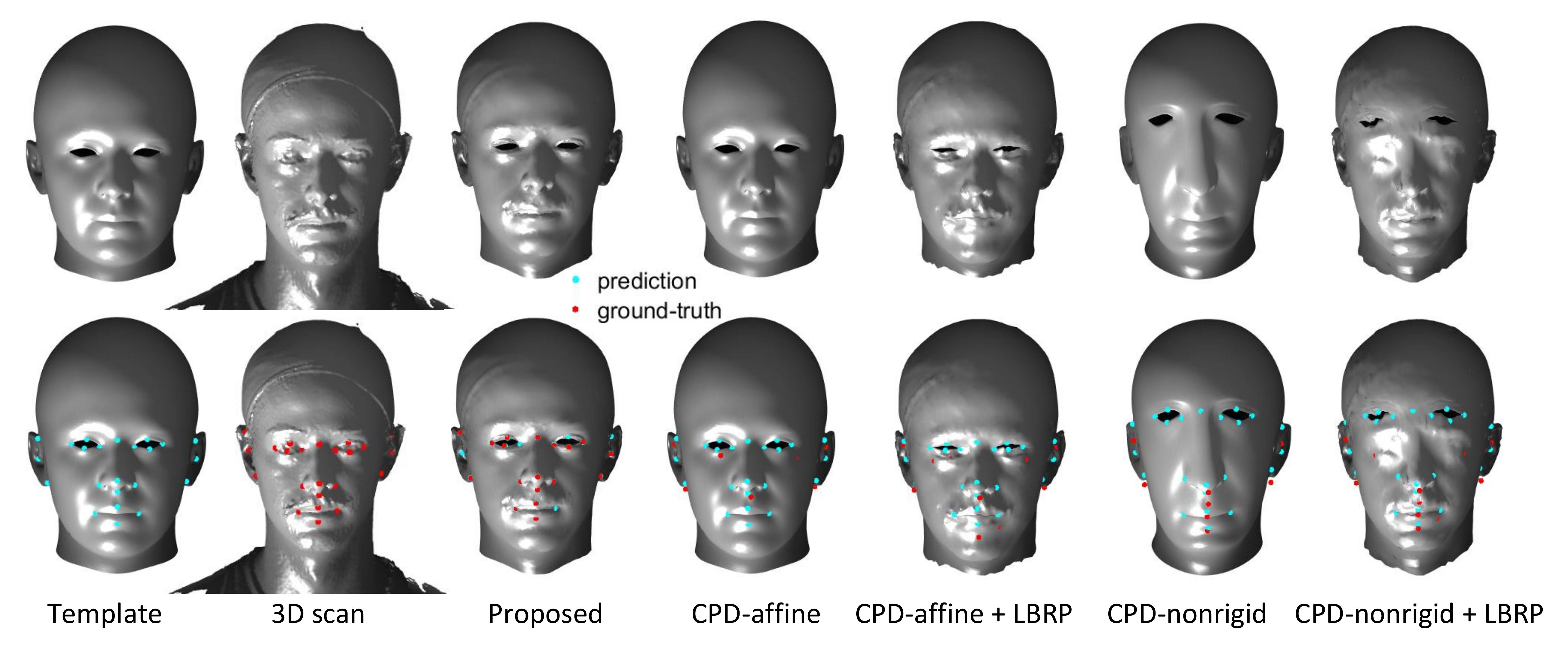}
\end{center}
   \caption{Proposed method compared with standard CPD. Ground truth points on target raw 3D data shown in red, corresponding template points shown in cyan.}
\label{fig:compareCPD}
\end{figure}

We present a new pipeline in fully-automatic non-rigid 3D shape registration by integrating several powerful ideas from the computer vision and graphics. These include Mixture-of-Trees 2D landmarking \cite{Ramanan12}, Iterative Closest Points (ICP) \cite{besl1992method}, Coherent Point Drift (CPD) \cite{myronenko2010}, and mesh editing using the Laplace-Beltrami (LB) operator \cite{Sorkine2004}. We also provide comparisons of the latter approach with the use of Gaussian Processes (GPs) \cite{Thomas17}.
Our contributions include: 1) a new 3D landmarking and pose normalisation method; 2) an adaptive shape template method to accelerate the convergence of registration algorithms and achieve a better final shape correspondence; 3)  a new iterative registration method that combines ICP with CPD to achieve a more stable and accurate correspondence establishment than standard CPD. We call this approach \emph{Iterative Coherent Point Drift} (ICPD). 

Our proposed pipeline is evaluated qualitatively and quantitatively on three human face/head datasets: Headspace \cite{headspace10,robertson2017morphable,Dai_2017_ICCV}, BU3D \cite{yin2008high} and LSFM-synthetic \cite{booth2018large}, and is compared with several other methods. Note that this latter dataset are 100 faces generated randomly from the Large Scale Face Model \cite{booth2018large} where samples lie within $\pm 3$SDs.The pipelineel achieves state-of-the-art performance. Fig. \ref{fig:compareCPD} is a qualitative illustration of a typical result where our method achieves a more accurate correspondence than standard CPD. Note that the landmarks on our method are almost exactly the same position as their corresponding ground-truth points on the raw 3D scan. Even though standard CPD-affine is aided by Laplace-Beltrami regularised projection (LBRP, a component of our proposed pipeline), the result shows a "squeezed" face around the eye and mouth regions and the landmarks are far away from their corresponding ground-truth positions. 

The rest of the paper is structured as follows. After presenting related work, we give a technical background on our template adaptation approach. In Sec. \ref{sec:framework} we describe our non-rigid registration framework, while the following section evaluates it over three datasets. Lastly we present conclusions.

\section{Related work}
\label{sec:lit}

Here we provide background literature to our pipeline, in the order in which the processes are required.

\subsection{Data to template alignment}

All template morphing methods need to align the raw data to be sufficiently close to the template, which is in some canonical position (eg frontal for a human face/head). Thus the template is brought within the convergence basin of the global minimum of alignment and morphing.
To this end, we use a 2D landmarker and project the detected landmarks to 3D.
In particular, Zhu and Ramanan \cite{Ramanan12} use a \emph{Mixture of Trees} model of the face, which both detects faces and locates facial landmarks. One of the major advantages of their approach is that it can handle extreme head poses even at relatively low image resolutions.

\subsection{Template adaptation}

By template adaptation, we mean the ability to adapt the shape of a template using \emph{sparse} shape information before applying \emph{dense} morphing. Here we overview the two methods evaluated in our pipeline: (i) Laplace-Beltrami mesh manipulation and (ii) the posterior model (PM) of a Gaussian Process Morphable Model (GPMM). 

\subsubsection{Laplace-Beltrami mesh manipulation}

The Laplace-Beltrami (LB) operator is widely used in 3D mesh manipulation. The LB term regularises the landmark-guided template adaptation in two ways: 1) the landmarks on the template are manipulated towards their corresponding landmarks on the raw 3D scan; 2) all other points in original template are moved \emph{As Rigidly As Possible} (ARAP \cite{Sorkine2007}) regarding the landmarks' movement, according to an optimised cost function, described later. Following Sorkine et al. \cite{Sorkine2007}, 
the idea for measuring the rigidity of a deformation of the whole mesh is to sum up over the deviations from rigidity. Thus, the energy functional can be formed as:
\begin{equation}       
\mathbf{E(S^{\prime})}  =  \sum_{i=1}^n \mathbf{w}_i  \sum_{j \in N(i)} \mathbf{w}_{ij}
\lVert (\mathbf{p^\prime}_i-\mathbf{p^\prime}_j) - \mathbf{R}_i(\mathbf{p}_i-\mathbf{p}_j)  \rVert,
\end{equation}
where we denote a mesh by $\mathbf{S}$, with $\mathbf{S^{\prime}}$ its deformed mesh and $\mathbf{R}$ is a rotation. Mesh topology is determined by $n$ vertices and $m$ triangles. Also $N(i)$ is the set of vertices connected to vertex $i$; these are the one-ring neighbours. The parameters $\mathbf{w}_i$, $\mathbf{w}_{ij}$ are fixed cell and edge weights. Note that $\mathbf{E(S^{\prime})}$ depends solely on the geometries of $\mathbf{S}$, $\mathbf{S^{\prime}}$, i.e., on the vertex positions $\mathbf{p}$, $\mathbf{p^\prime}$. In particular, since the reference mesh (our input shape) is fixed, the only variables in $\mathbf{E(S^{\prime})}$ are the deformed vertex positions $\mathbf{p^\prime}_i$. The gradient of $\mathbf{E(S^{\prime})}$ is computed with respect to
the positions $\mathbf{p^\prime}$. The partial derivatives w.r.t. $\mathbf{p^\prime}_i$ can be computed as:
\begin{equation}       
\dfrac{d \mathbf{E(S^{\prime})}}{d \mathbf{p^\prime}_i}  =  \sum_{j \in N(i)} 4\mathbf{w}_{ij}
\left( (\mathbf{p^\prime}_i-\mathbf{p^\prime}_j) - \frac{1}{2} (\mathbf{R}_i + \mathbf{R}_j) (\mathbf{p}_i-\mathbf{p}_j)  \right)
\end{equation}
Setting the partial derivatives to zero w.r.t. each $\mathbf{p^\prime}_i$ gives the following sparse linear system of equations:
\begin{equation}       
\sum_{j \in N(i)} \mathbf{w}_{ij} (\mathbf{p^\prime}_i-\mathbf{p^\prime}_j)  =  \sum_{j \in N(i)} \frac{\mathbf{w}_{ij}}{2} (\mathbf{R}_i + \mathbf{R}_j) (\mathbf{p}_i-\mathbf{p}_j)
\label{eqt:pderivatives2zero}
\end{equation}
The linear combination on the left-hand side is the discrete Laplace-Beltrami operator applied to $\mathbf{p^\prime}$, hence the system of equations can be written as:
\begin{equation}       
\mathbf{L}\mathbf{p^\prime}  =  \mathbf{b},
\label{eqt:LBlinear}
\end{equation}
where $\mathbf{b}$ is an n-vector whose i-th row contains the right-hand side expression from (\ref{eqt:pderivatives2zero}). We also need to incorporate the
modeling constraints into this system. In the simplest form, those can be expressed by some fixed positions
\begin{equation}       
\mathbf{p^\prime}_j  =  \mathbf{c}_k, k \in \mathcal{F},
\end{equation}
where $\mathcal{F}$ is the set of indices of the constrained vertices. 
In our case, these are the landmark positions, automatically detected on the raw 3D data, with the corresponding points known \emph{a priori} on the template.
Incorporating such constraints into (\ref{eqt:LBlinear}) requires substituting the corresponding variables, erasing respective rows and columns from $\mathbf{L}$ and updating the right-hand side with the values $\mathbf{c}_k$.

\subsubsection{Gaussian process morphable model}

A Gaussian Process Morphable Model (GPMM) uses manually defined arbitrary kernel functions to describe the deformation's covariance matrix. This enables a GPMM to aid the construction of a 3DMM, without the need for training data. The posterior models (PMs) of GPMMs are regression models of the shape deformation field. Given partial observations, such posterior models are able to determine what is the potential complete shape. A posterior model is able to estimate other points' movements when some set of landmarks and their target positions are given.

Instead of modelling absolute vertex positions using PCA, GPMMs model a shape as a deformation vector field $\mathbf{u}$ from a reference shape $\mathbf{X} \in \mathbb{R}^{p \times 3}$, i.e. a shape $\mathbf{X^\prime}$ can be represented as
\begin{equation}       
\mathbf{X^\prime}  = \mathbf{X} + \mathbf{u}(\mathbf{X})
\end{equation}
for some deformation vector field $\mathbf{u} \in \mathbb{R}^{p \times 3}$. We model the deformation as a Gaussian process $\mathbf{u} \sim GP(\mathbf{\mu}, \mathbf{k})$ where $\mathbf{\mu} \in \mathbb{R}^{p \times 3}$ is a mean deformation and $\mathbf{k} \in \mathbb{R}^{3 \times 3}$ a covariance function or kernel.

The biggest advantage of GPMMs compared to statistical shape models (eg 3DMMs) is that we have much more freedom in defining the covariance function. GPMMs allow expressive prior models for registration to be derived, by leveraging the modeling power of Gaussian processes. By estimating the covariances from example data GPMMs becomes a continuous version of a statistical shape model.

\subsection{Dense shape registration}

The Iterative Closest Points (ICP) algorithm \cite{arun1987least,besl1992method} is the standard rigid-motion registration method. 
Several extensions of ICP for the nonrigid case were proposed \cite{Amberg07,booth2018large,hontani2012robust,cheng2015active,cheng2017statistical,kou2016modified}. Often these have good performance in shape difference elimination but have problems in over fitting and point sliding.
Another approach is based on modelling the transformation with thin plate splines (TPS) \cite{bookstein1989principal} followed by robust point matching (RPM) and is known as TPS-RPM \cite{chui2000new}. However, it is slow in large-scale point set registration \cite{yang2011thin,lee2011topology,lee2015non,ma2017non}. 
Amberg et al. \cite{Amberg07} defined the optimal-step \emph{Nonrigid Iterative Closest Points} (NICP) framework.
Recently Booth et al.~\cite{booth2018large} built a Large Scale Facial Model (LSFM), using the same NICP template morphing approach with error pruning, followed by Generalised Procrustes Analysis (GPA) for alignment, and Principal Component Analysis (PCA) for the model construction. Li et al. \cite{li2008global} show that using proximity heuristics to determine correspondences is less reliable when large deformations are present. Their Global Correspondence Optimization approach solves simultaneously for both the deformation parameters and correspondences \cite{li2008global}. 

Myronenko et al. consider the alignment of two point sets as a probability density estimation \cite{myronenko2010} and they call the method Coherent Point Drift (CPD). There is no closed-form solution for this optimisation, so it employs an EM algorithm to optimize the Gaussian Mixture Model (GMM) fitting. Algorithms are provided to solve for several shape deformation models such a affine (CPD-affine) and generally non-rigid (CPD-nonrigid). Their \emph{`non-rigid'} motion model employs an $M \times M$ Gaussian kernel ${\bf G}$ for motion field smoothing, and the M-step requires solving for an $M \times3$ matrix ${\bf W}$ that generates the template deformation (GMM motion field) as ${\bf GW}$.  Such motion regularisation is related to motion coherence, and inspired the algorithm's name. The CPD method was has been extended by various groups \cite{wang2011refined,golyanik2016extended,hu2010deformable,trimech20173d}. 
Compared to TPS-RPM, CPD offers superior accuracy and stability with respect to non-rigid deformations in presence of outliers. A modified version of CPD imposed a \emph{Local Linear Embedding} topological constraint to cope with highly articulated non-rigid deformations \cite{ge2014non}. However, this extension is more sensitive to noise than CPD. A non-rigid registration method used Student’s Mixture Model (SMM) to do probability density estimation \cite{zhou2014robust}. The results are more robust and accurate on noisy data than CPD. Dai et al. \cite{Dai_2017_ICCV} proposed a hierarchical parts-based CPD-LB morphing framework to avoid under-fitting and over-fitting. It overcomes the sliding problem to some extent, but the end result still has a small tangential error.

Marcel et al.~\cite{luthi2017gaussian} model shape variations with a Gaussian process (GP), which they represent using the leading components of its Karhunen-Loeve expansion. Such Gaussian Process Morphable Models (GPMMs) unify a variety of non-rigid deformation models. 
Gerig et al. \cite{Thomas17} present a novel pipeline for morphable face model construction based on Gaussian processes. GPMMs separate problem specific requirements from the registration algorithm by incorporating domain-specific adaptions as a prior model.

\section{Non-rigid shape registration framework}
\label{sec:framework}
\begin{figure}[t!]
\begin{center}
\includegraphics[clip, trim=0.5cm 0.5cm 0.5cm 0.5cm,width=1\linewidth]{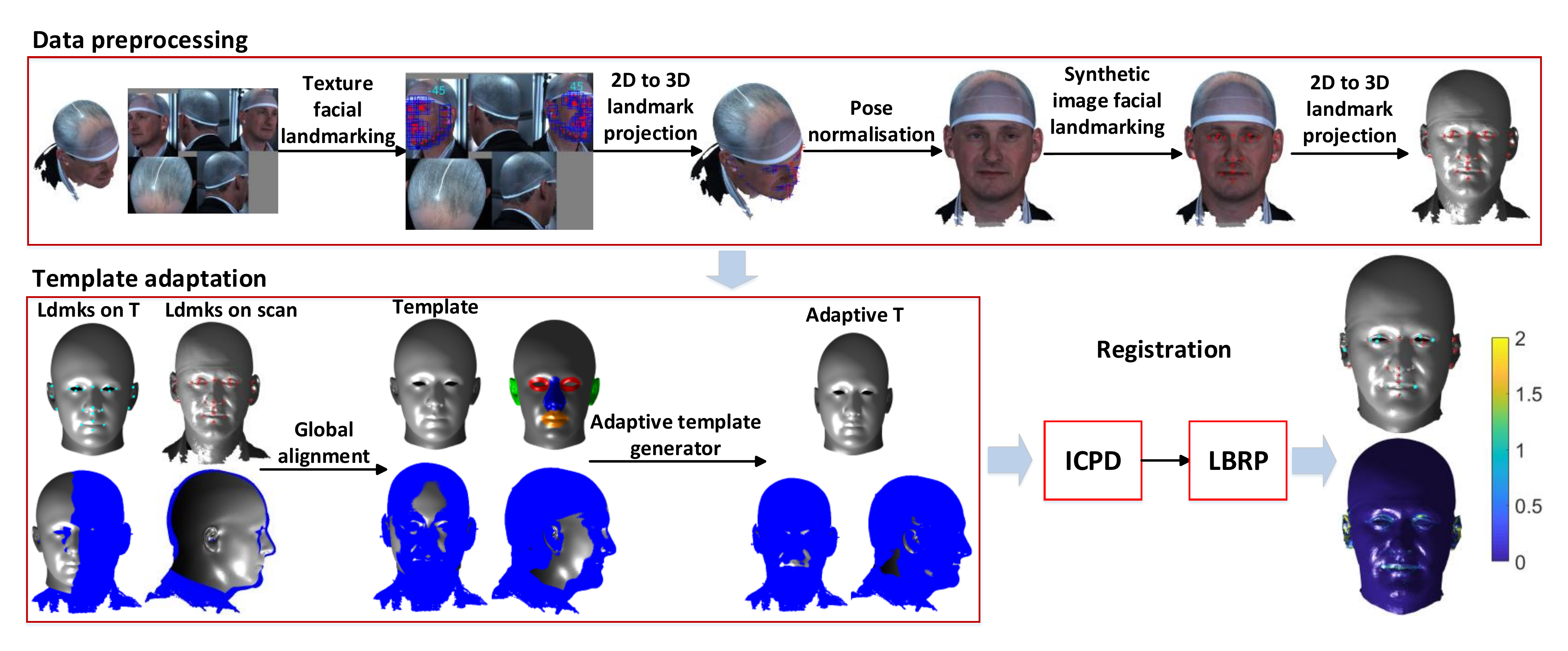}
\end{center}
   \caption{Adaptive template registration framework, using ICPD based morphing.}
\label{fig:flow}
\end{figure}

The proposed registration framework is shown in Fig. \ref{fig:flow} and includes four high-level stages: 1) data preprocessing of raw 3D image: landmarking and pose normalisation; 2) template adaptation: global alignment and adapting the template shape; 3) template morphing using Iterative Coherent Point Drift (ICPD); 4) point projection regularised by the Laplace-Beltrami operator (LBRP). These are detailed in the following four subsections.

\subsection{Data preprocessing}

Data preprocessing of the raw 3D scan serves to place the data in a frontal pose, which allows us to get a complete and accurate set of automatic 3D landmark positions, for every 3D image, that correspond to a set of manually-placed (once) landmarks on the template. This preprocessing comprises five sub-stages: (i) 2D landmarking, (ii) projection to 3D landmarks, (iii) pose normalisation (iv) synthetic frontal 2D image landmarking and (v) projection to 3D landmarks.

We use the `Mixture of Trees' method of Zhu and Ramanan \cite{Ramanan12} to localise 2D facial landmarks. In particular, the mixture we use has 13 landmark tree models for 13 different yaw angles of the head.

We apply the detector to the composite 2D image that contains all 5 viewpoints of the capture system, see \ref{fig:flow}, top left. Two face detections are found, of approximately 15 degrees and 45 degrees yaw from the frontal pose, corresponding to the left and right side of the face respectively. The detected 2D landmarks are then projected to 3D using the OBJ texture coordinates in the raw data.

Given that we know where all of these 3D landmarks should be for a frontal pose, it is possible to do standard 3D pose alignment in a scale-normalised setting \cite{Dai_2017_ICCV}.


In around 1\% of the dataset, only one tree is detected and that is used for pose normalisation, and in the rest 2-3 images are detected. In the cases where 3 trees are detected, the lowest scoring tree is always false positive and can be discarded. For the remaining two trees, a weighted combination of the two rotations is computed using quaternions, where the weighting is based on the  mean Euclidean error to the mean tree, in the appropriate tree component.

After we have rotated the 3D image to canonical frontal view, we wish to generate a set of landmarks that are accurate and correspond to the set marked up on the template. This is the set related to the central tree (0 degrees yaw) in the mixture. After these 2D landmarks are extracted, they are again projected to 3D  using the raw OBJ texture coordinates.

\subsection{Template adaptation}

As shown in Fig. \ref{fig:flow}, template adaptation consists of two sub-stages: (i) global alignment followed by (ii) dynamically adapting the template shape to the data. 
For global alignment, we manually select the same landmarks on the template as we automatically extract on the raw data (i.e. using the zero yaw angle tree component from \cite{Ramanan12}). Note that this needs to be done \emph{once only} for some object class and so doesn't impact on the autonomy of the \emph{online} operation of the framework. Then we align rigidly (without scaling) from the 3D landmarks on raw 3D data to the same landmarks on the template. The rigid transformation matrix is used for the raw data alignment to the template.

The template is then \emph{adapted} to better align with the raw scan. A better template helps the later registration converge faster and gives more accurate correspondence at the beginning and end of registration. A good template has the same size and position of local facial parts (e.g. eyes, nose, mouth and ears) as the raw scan. This cannot be achieved by mesh alignment alone. We propose two method to give a better template that is adapted to the raw 3D scan: (1) Laplace-Beltrami mesh editing; (2) Template estimation via posterior GPMMs. For both methods, three ingredients are needed: landmarks on 3D raw data, the corresponding landmarks on template, and the original template. 

\subsubsection{Laplace-Beltrami mesh manipulation:}
We decompose the template into several facial parts: eyes, nose, mouth, left ear and right ear. We rigidly align landmarks on each part separately to their corresponding landmarks on 3D raw data. These rigid transformation matrices are used for aligning the decomposed parts to 3D raw data. The rigidly transformed facial parts tell the original template where it should be. We treat this as a mesh manipulation problem. We use Laplace-Beltrami mesh editing to manipulate the original template towards the rigidly transformed facial parts, as follows: (1) the \emph{facial parts} (fp) of the original template are manipulated towards their target positions - these are rigidly transformed facial parts; (2) all other parts of the original template are moved \emph{as rigidly as possible} \cite{Sorkine2007}.

Given the vertices of a template stored in the matrix $\mathbf{X}_{\textrm{T}} \in \mathbb{R}^{p \times 3}$ and a better template obtained whose vertices are stored in the matrix $\mathbf{X}_{\textrm{bT}} \in \mathbb{R}^{p \times 3}$, we define the selection matrices $\mathbf{S}_{fp}\in [0,1]^{l \times p}$ as those that select the $l$ vertices (facial parts in $\mathbf{X}_{\textrm{T}}$ and $\mathbf{X}_{\textrm{bT}}$) from the raw template and a better template respectively. This linear system can be written as:
\begin{equation}       
\left(                 
  \begin{array}{c}   
    \mathbf{\lambda L}\\ 
    \quad \mathbf{S}_{fp}\\  
  \end{array}
\right)\mathbf{X}_{\textrm{bT}}  =   
\left(                 
  \begin{array}{c}   
   \quad \mathbf{\lambda L} \mathbf{X}_{\textrm{T}}\\ 
    \mathbf{X}_{\textrm{fp}}\\  
  \end{array}
\right)
\end{equation}
where $\mathbf{L}\in \mathbb{R}^{p \times p}$ is the cotangent Laplacian approximation to the LB operator and $\mathbf{X}_{\textrm{bT}}$ is the better template that we wish to solve for. The parameter $\lambda$ weights the relative influence of the position and regularisation constraints, effectively determining the `stiffness' of the mesh manipulation. As $\lambda \rightarrow 0$, the facial parts of the original template are manipulated exactly to the rigidly transformed facial parts. As $\lambda \rightarrow \infty$, the adaptive template will only be at the same position as the original template $\mathbf{X}_{\textrm{T}}$.

\begin{figure}[t!]
\begin{center}
\includegraphics[clip, trim=1cm 0.7cm 1cm 0.7cm,width=1\linewidth]{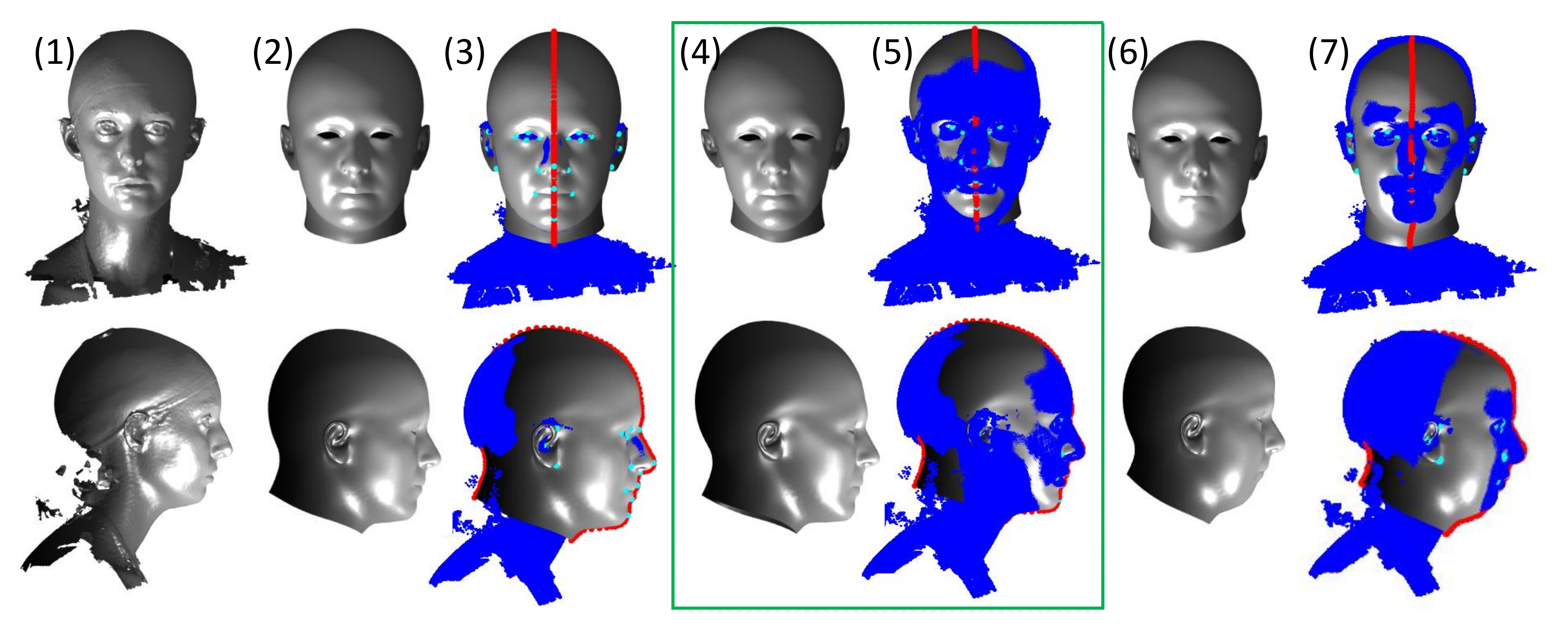}
\end{center}
   \caption{Template comparison with raw data: (1) raw scan; (2) template with global rigid alignment; (3) 2 compared with the raw scan; (4) adaptive template via LB mesh editing; (5) 4 compared with the raw scan; (6) the mean template estimation via posterior models;
   (7) 6 compared with the raw scan in (7)}
\label{fig:templateCompare}
\end{figure}

\subsubsection{Template estimation via posterior models:}
A common task in shape modelling is to infer the full shape from a set of measurements of the shape. This task can be formalised as a regression problem. The posterior models of Gaussian Process Morphable Models (GPMMs) are regression models of the deformation field. Given partial observations, posterior models are able to answer what is the potential full shape. Posterior models show the points' potential movements when the landmarks are fixed to their target position.

In a GPMM, let $\left\{ x_1,...,x_l \right\} \in \mathbb{R}^{l \times 3}$ be a fixed set of input 3D points and assume that there is a regression function $f_0\rightarrow \mathbb{R}^{p \times 3}$, which generates a new vector field $\mathbf{y_i} \in \mathbb{R}^{p \times 3}$ according to
\begin{equation}  
\mathbf{y_i} = f_0(\mathbf{x_i})+\mathbf{\epsilon_i},(i=1,...,n).
\end{equation}
where $\mathbf{\epsilon_i}$ is independent Gaussian noise, i.e. $\mathbf{\epsilon_i} \sim N(0,\delta^2)$. The regression problem is to infer the function $f_0$ at the input points $\left\{ x_1,...,x_l \right\}$. The possible deformation field $\mathbf{y_i}$ is modelled using a Gaussian process model $GP(\mu,k)$ that models the shape variations of a given shape family. 

In our case, the reference shape is the original template, the landmarks on the original template are the fixed set of input 3D points. The same landmarks on 3D raw data are the target position of the fixed set of input 3D points. Given a GPMM $GP(\mu,k)$ that models the shape variations of a shape family, the adaptive template is 
\begin{equation}  
\mathbf{X}^{i}_{\textrm{bT}} =\mathbf{X}_{\textrm{T}} + \mathbf{y_i},(i=1,...,n).
\end{equation}
The mean of $\mathbf{X}^{i}_{\textrm{bT}}$ is shown in Fig. \ref{fig:templateCompare} (6) and (7).

\subsection{Iterative coherent point drift}
\label{sec:ICPD}

The task of non-rigid 3D registration (shape morphing) is to deform and align the template to the target raw 3D scan. 
Non-rigid Coherent Point Drift (CPD) \cite{myronenko2010} has better deformation results when partial correspondences are given and we have found that it is more stable and converges better when the template and the raw data have approximately the same number of points. However, the correspondence is often not known before registration. Thus, following an Iterative Closest Points (ICP) scheme \cite{besl1992method}, we supply CPD registration with coarse correspondences using `closest points'. We refine such correspondences throughout iterations of the \emph{Iterative Coherent Point Drift} (ICPD) approach described here. 

We use the original code package of CPD available online as library calls for ICPD. Other option parameters can be found in the CPD author's release code. The global affine transformation is used as a small adjustment of correspondence computation. A better correspondence (idx2 in the pseudocode) is used as the priors for CPD non-rigid registration. The qualitative output of ICPD is very smooth, a feature inherited from standard CPD. A subsequent regularised point projection process is required to capture the target shape detail, and this is described next. 

\subsection{Laplace-Beltrami regularised projection}
\label{sec:LBRP}
When ICPD has deformed the template close to the scan, point projection is required to eliminate any (normal) shape distance error. 
Again, we overcome this by treating the projection operation as a mesh editing problem with two ingredients. First, position constraints are provided by those vertices with mutual nearest neighbours between the deformed template and raw scan. Using mutual nearest neighbours reduces sensitivity to missing data. Second, regularisation constraints are provided by the LB operator which acts to retain the local structure of the mesh. We call this process \emph{Laplace-Beltrami regularised projection} (LBRP), as shown in the registration framework in Fig. \ref{fig:flow}.

We write the point projection problem as a linear system of equations. 
Given the vertices of a scan stored in the matrix $\mathbf{X}_{\textrm{scan}} \in \mathbb{R}^{n \times 3}$ and the deformed template obtained by CPD whose vertices are stored in the matrix $\mathbf{X}_{\textrm{deformed}} \in \mathbb{R}^{p \times 3}$, we define the selection matrices $\mathbf{S}_{1}\in [0,1]^{m \times p}$ and $\mathbf{S}_{2}\in [0,1]^{m \times n}$ as those that select the $m$ vertices with mutual nearest neighbours from deformed template and scan respectively. This linear system can be written as:
\begin{equation}       
\left(                 
  \begin{array}{c}   
    \mathbf{\lambda L}\\ 
    \quad \mathbf{S}_{1}\\  
  \end{array}
\right)\mathbf{X}_{\textrm{proj}}  =   
\left(                 
  \begin{array}{c}   
   \quad \mathbf{\lambda L} \mathbf{X}_{\textrm{deformed}}\\ 
    \mathbf{S}_{2} \mathbf{X}_{\textrm{scan}}\\  
  \end{array}
\right)
\end{equation}
where $\mathbf{L}\in \mathbb{R}^{p \times p}$ is the cotangent Laplacian approximation to the LB operator and $\mathbf{X}_{\textrm{proj}} \in \mathbb{R}^{p \times 3}$ are the projected vertex positions that we wish to solve for. The parameter $\lambda$ weights the relative influence of the position and regularisation constraints, effectively determining the `stiffness' of the projection. As $\lambda \rightarrow 0$, the projection tends towards nearest neighbour projection. As $\lambda \rightarrow \infty$, the deformed template will only be allowed to rigidly transform.

\section{Evaluation}
\label{sec:evaluation}
We evaluated the proposed registration framework using three datasets: Headspace \cite{headspace10,robertson2017morphable,Dai_2017_ICCV}, BU3D (neutral expression) \cite{yin2008high} and LSFM-synthetic. The latter two have ground-truth information. We use error to manually-defined landmark and the average nearest point distance error for evaluation on the Headspace dataset. Recently, two registration frameworks have become publicly available for comparison: Basel's Open Framework (OF) \cite{luthi2017gaussian} and the LSFM pipeline \cite{booth2018large}.

For reproducibility of our results, parameters used are as follows: 1) when doing the 3D automatic face landmarking, we manually select the same landmarks on template mesh only for once across the whole dataset; 2) $\mathbf{\lambda}$ is set to 0.1; 3) the iteration limitation of CPD-affine is 200 and CPD-non-rigid is 300. 4) the Gaussian kernels in GPMM are defined as the same as that in \cite{Thomas17}.

\subsection{Internal comparison of approaches}
\begin{figure}[t!]
\begin{center}
\includegraphics[clip, trim=0.5cm 1.5cm 0.5cm 0.6cm,width=1\linewidth]{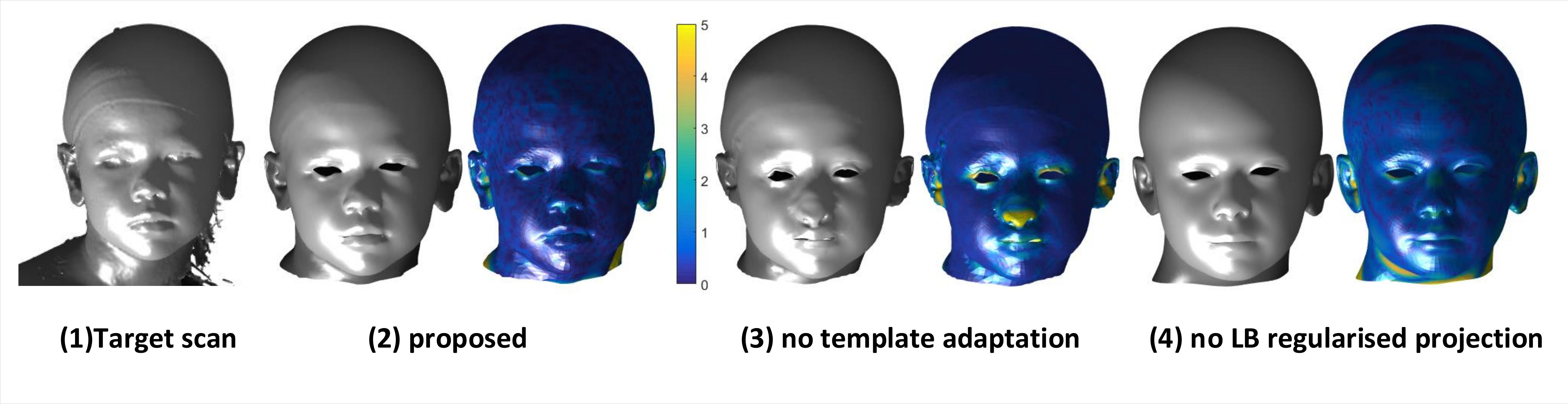}
\end{center}
   \caption{(1) target scan; (2) proposed method (3) remove template adaptation; (4) remove LB regularised projection. Error map (mm).}
\label{fig:selfcompare}
\end{figure}

We validate the effectiveness of each step in the proposed registration pipeline qualitatively and quantitatively. The results of registration over children data in Headspace are shown in Fig.\ref{fig:selfcompare}. After pure rigid alignment without template adaptation, the nose of template is still bigger than the target. As can be seen in Fig. \ref{fig:selfcompare} (3), the nose ends up with a bad deformation result. The same problem happened in the ear. Without LB regularised projection shown in Fig. \ref{fig:selfcompare} (4), it fails in capturing the shape detail compared with the proposed method.

\begin{figure}[t!]
\begin{center}
\includegraphics[clip, trim=0.5cm 0.5cm 0.5cm 0.5cm,width=0.9\linewidth]{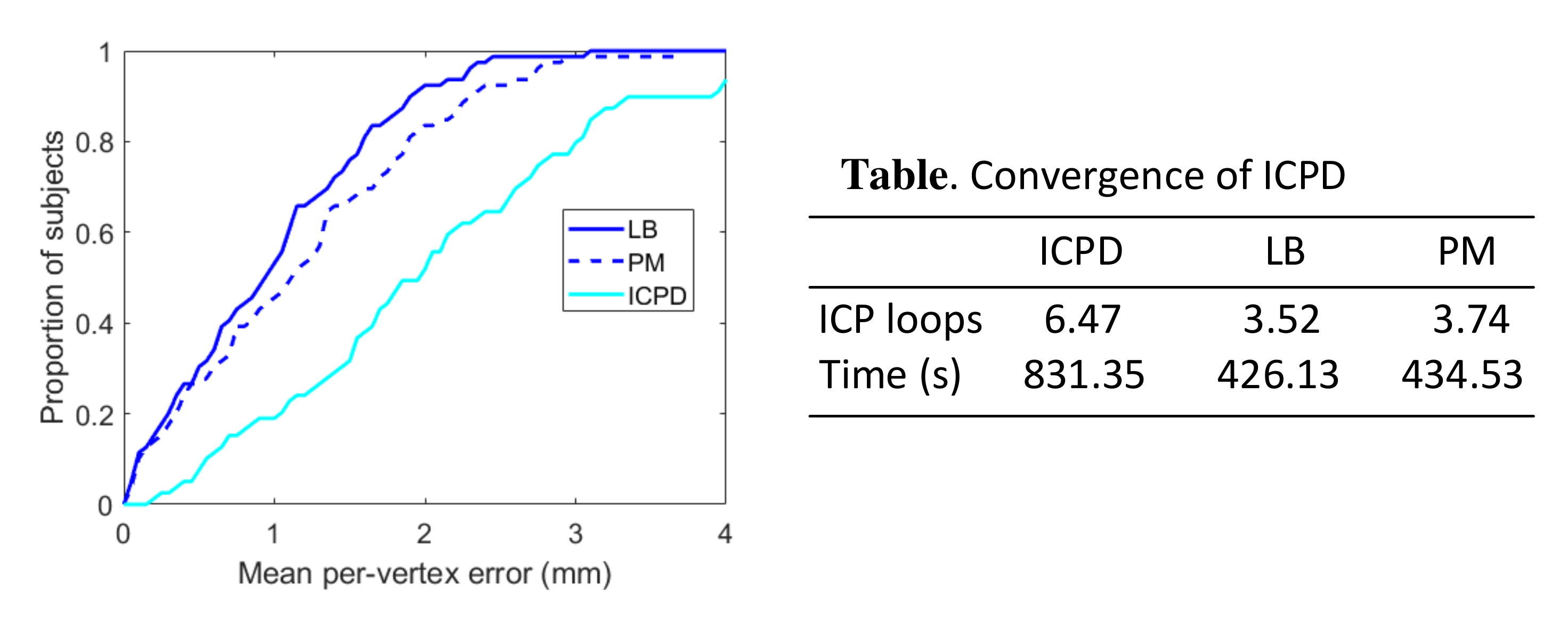}
\end{center}
   \caption{Improvement in correspondence and convergence performance when using adaptive templates: 1) ICPD without an adaptive template (cyan) ; 2) ICPD with LB-based adaptive template (blue); 3).ICPD with adaptive PM-based template (blue dashed).}
\label{fig:convergence}
\end{figure}

Using the BU3D dataset for quantitative validation, we compared the performance of (i) the proposed ICPD registration, (ii) ICPD with an adaptive template using LB mesh manipulation and (iii) ICPD with an adaptive template, using a posterior model (PM). The mean per-vertex error is computed between the registration results and their ground-truth. The number of ICPD iterations and computation time is recorded, when using the same computation platform. The per-vertex error plot in Fig. \ref{fig:convergence} illustrates that the adaptive template improves the correspondence accuracy of ICPD. The number of ICPD iterations and computation time is significantly decreased by the adaptive template method. In particular adaptive template using LB mesh manipulation has better performance than adaptive template using a posterior model. Thus, we employ an adaptive template approach using LB mesh manipulation for later experiments.    

\subsection{Correspondence comparison}

\subsubsection{Headspace:}
\begin{figure}[t!]
\begin{center}
\includegraphics[clip, trim=1cm 0.5cm 1cm 0.5cm,width=0.8\linewidth]{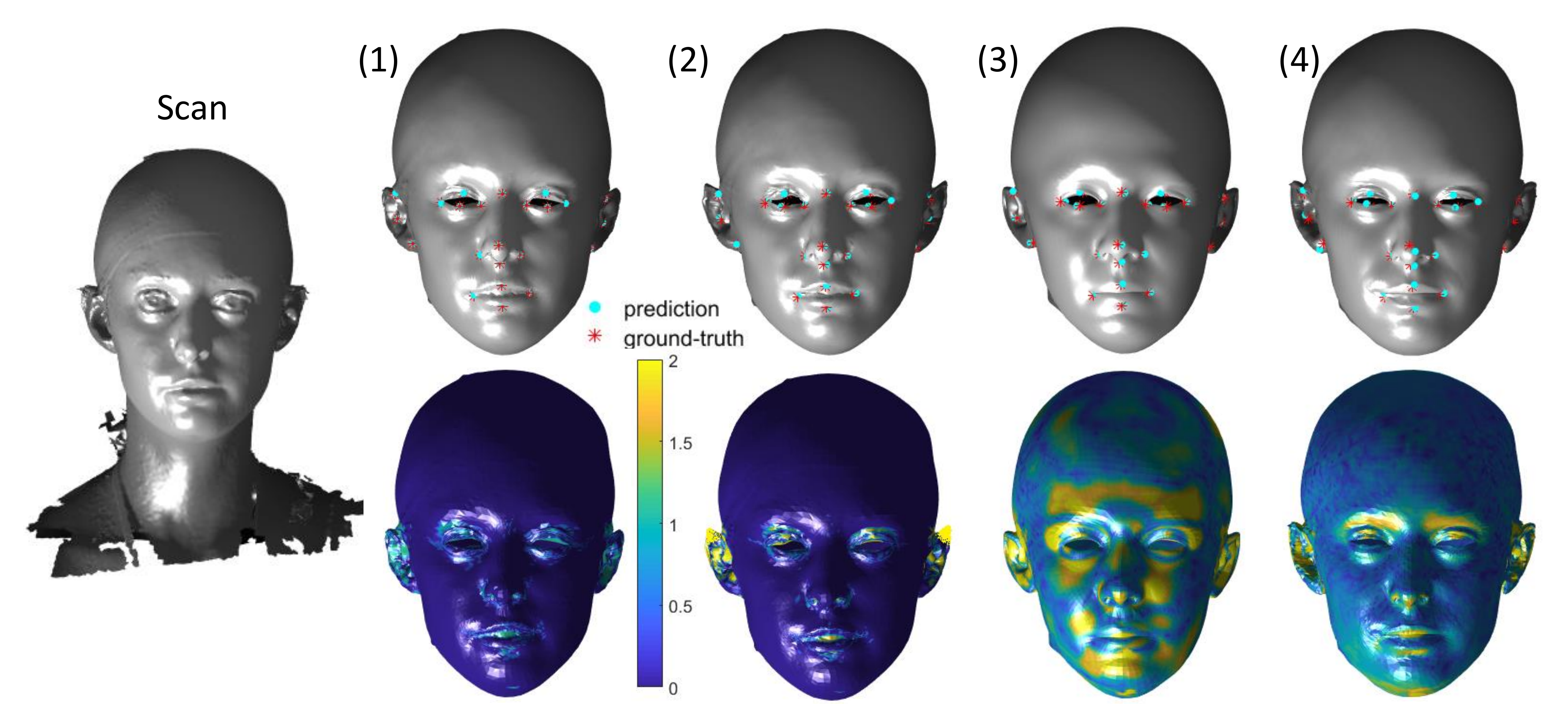}
\end{center}
   \caption{First row - correspondence results and their landmarks compared with ground-truth on raw scan; Second row - the color map of per-vertex nearest point error. (1) proposed method with LB template adaptation; (2) proposed method without adaptive template; (3) Open Framework morphing \cite{luthi2017gaussian}; (4) LSFM morphing \cite{booth2018large}.}
\label{fig:correspondenceComapre}
\end{figure}

\begin{figure}[t!]
\begin{center}
\includegraphics[width=0.45\linewidth]{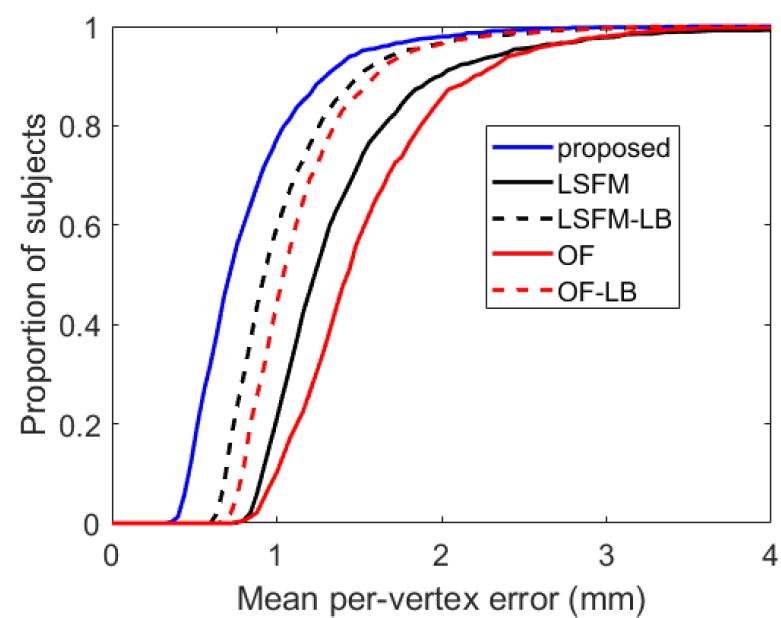}
\end{center}
   \caption{Mean per-vertex in morphed template nearest point distance error, higher is better.}
\label{fig:correspondenceHead}
\end{figure}

We evaluate correspondence accuracy both qualitatively and quantitatively. 1212 scans from Headspace are used for evaluation. A typical registration result is shown in Fig. \ref{fig:correspondenceComapre}. Apart from the proposed method, there are clearly significant errors around the ear region, the eye region, or even multiple regions. We use per-vertex nearest point error for quantitative shape registration evaluation. The per-vertex nearest point error is computed by measuring the nearest point distance from the morphed template to raw scan and averaging over all vertices. As can be seen in Fig. \ref{fig:correspondenceHead}, the proposed method has the best performance, when compared to Basel's Open Framework (OF) \cite{luthi2017gaussian} and the LSFM \cite{booth2018large} registration approach.  The OF method has a smoothed output without much shape detail. The LSFM method captures shape detail, but it has greater landmark error and per-vertex nearest point error. The quantitative evaluation in Fig. \ref{fig:correspondenceHead} validates that the proposed method outperforms the other two contemporary methods.

\subsubsection{BU3D:}
\begin{figure}[t!]
\begin{center}
\includegraphics[clip, trim=1cm 0.5cm 1cm 0.5cm,width=0.8\linewidth]{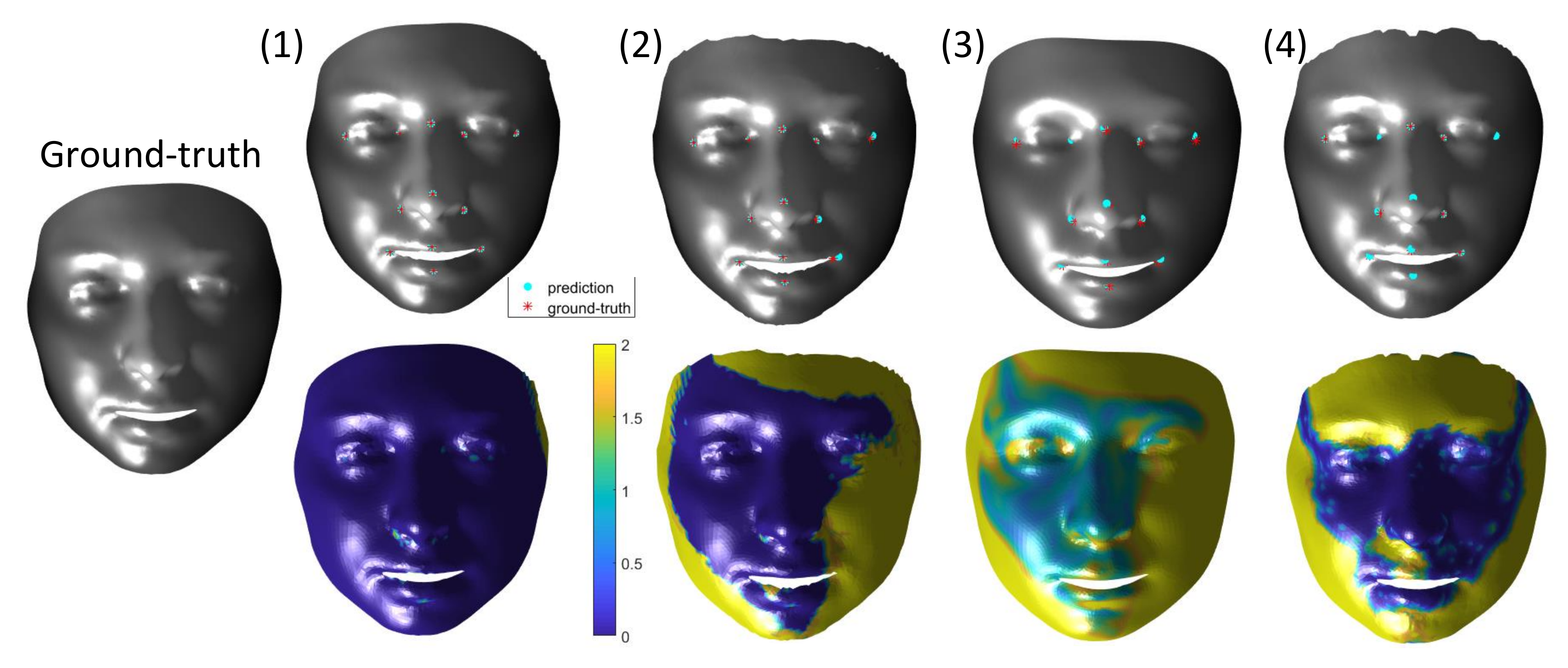}
\end{center}
   \caption{First row - correspondence results and their landmarks; Second row - the color map of per-vertex error against ground-truth. (1) proposed method; (2) proposed method without adaptive template; (3) OF registration; (4) LSFM registration.}
\label{fig:correspondenceComaprebu3d}
\end{figure}

\begin{figure}[t!]
\begin{center}
\includegraphics[clip, trim=0.5cm 0.5cm 0.5cm 0.5cm,width=0.8\linewidth]{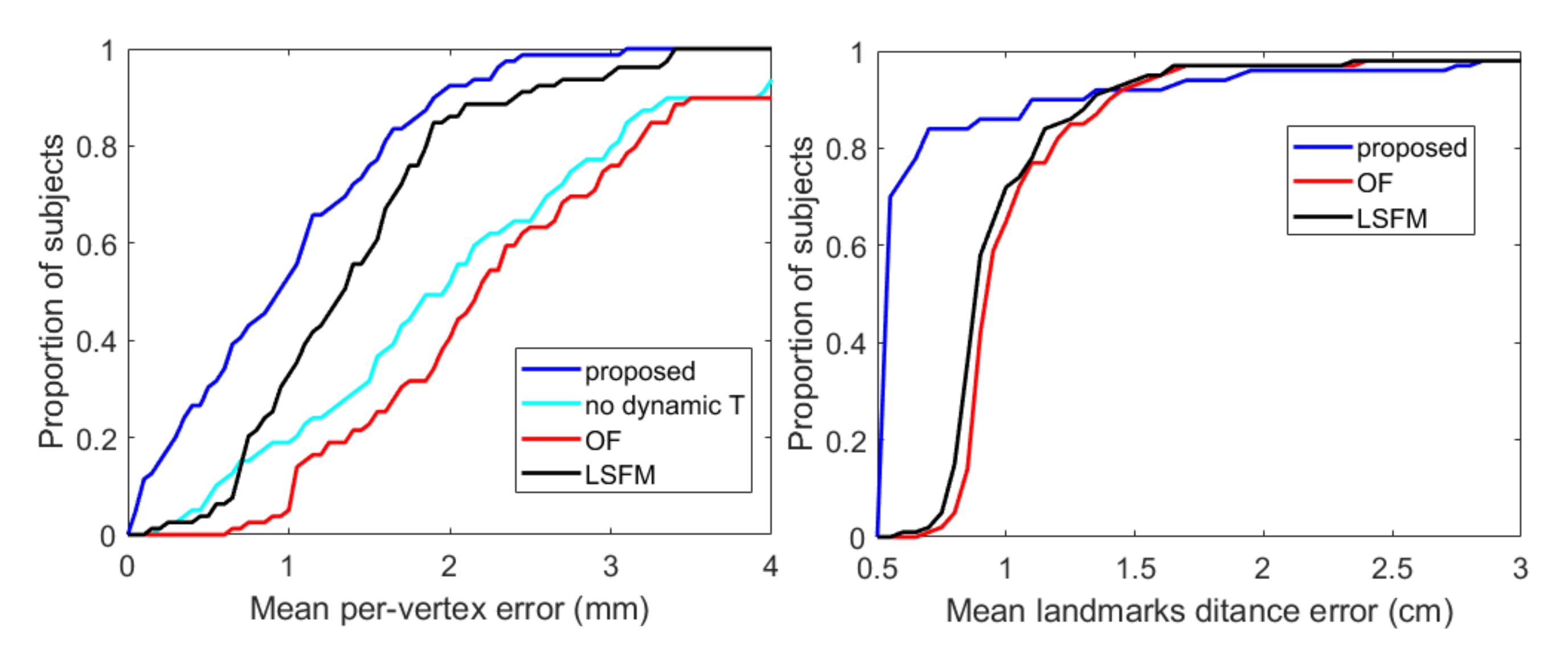}
\end{center}
   \caption{Mean per-vertex error: (1) left - BU3D dataset, (2) right - LSFM dataset }
\label{fig:correspondencefacebody}
\end{figure}

For the BU3D dataset, 100 scans with neutral expression are used for evaluation. We use 12 landmarks to perform adaptive template generation. Qualitatively from Fig. \ref{fig:correspondenceComaprebu3d} (1) and (2), the adaptive template can be seen to improve the registration performance. As shown in Fig. \ref{fig:correspondencefacebody} (1), compared with the ground-truth data, over 90\% of the registration results have less than 2 mm per-vertex error. The proposed method has the best performance in face shape registration. 

\subsubsection{LSFM:}
The synthetic LSFM dataset are 100 faces generated randomly from the Large Scale Face Model \cite{booth2018large} where samples lie within $\pm 3$SDs. The pipeline achieves state-of-the-art performance. we use the same 14 landmarks in \cite{Creusot2013} for the adaptive template generation. Since the synthetic data is already in correspondence, the 14 landmarks have the same indices across the dataset. As shown in Fig. \ref{fig:correspondencefacebody} (2), compared with the ground-truth data, over 83\% of the registration results have less than 1 mm per-vertex error. The proposed method has the best performance in synthetic data registration. 

\section{Conclusions}
\label{sec:conclusions}
We proposed a new fully-automatic shape registration framework with an adaptive template initialisation. Although there is a prior one-shot manual markup of landmarks on a generic template, this does not prevent our online process being fully-automatic.
The adaptive template accelerated the convergence of registration algorithms and achieved a more accurate correspondence. We provided two methods: LB mesh manipulation and the posterior model of the GPMM to achieve template adaptation. In particular, an adaptive template using LB mesh manipulation has a better performance than an adaptive template using a GP posterior model. We proposed a new morphing method that combined the ICP and CPD algorithms that is both more stable and accurate in correspondence establishment. We evaluated the proposed framework on three datasets: Headspace, BU3D and LSFM-synthetic. The proposed framework has better performance than other methods across all datasets.

\bibliographystyle{splncs}
\bibliography{egbib}
\end{document}